\documentclass[11pt,a4paper]{article}
\usepackage{fullpage}
\usepackage{authblk}
 
\usepackage[utf8]{inputenc} 
\usepackage[T1]{fontenc}    
\usepackage{hyperref}   
\usepackage{url}            
\usepackage{booktabs}      
\usepackage{nicefrac}      
\usepackage{microtype}
\usepackage{multirow}
\usepackage{multicol}

\usepackage{graphicx}
\usepackage{amsthm}
\usepackage{amssymb}
\usepackage{amsmath,bm,dsfont}
\usepackage{algorithmic,algorithm}

\title{Data-Efficient Fine-Tuning of Vision-Language Models for Diagnosis of Alzheimer’s Disease}

\author[1]{Fangqi Cheng}
\author[1]{Surajit Ray}
\author[1]{Xiaochen Yang\footnote{Corresponding author: Xiaochen Yang (email: xiaochen.yang@glasgow.ac.uk)}}
\affil[1]{School of Mathematics and Statistics, University of Glasgow, UK.}
\date{}

\begin{document}
\maketitle

\begin{abstract}

Medical vision-language models (Med-VLMs) have shown impressive results in tasks such as report generation and visual question answering, but they still face several limitations. Most notably, they underutilize patient metadata and lack integration of clinical diagnostic knowledge. Moreover, most existing models are typically trained from scratch or fine-tuned on large-scale 2D image-text pairs, requiring extensive computational resources, and their effectiveness on 3D medical imaging is often limited due to the absence of structural information. 
To address these gaps, we propose a data-efficient fine-tuning pipeline to adapt 3D CT-based Med-VLMs for 3D MRI and demonstrate its application in Alzheimer’s disease~(AD) diagnosis. Our system introduces two key innovations. 
First, we convert structured metadata into synthetic reports, enriching textual input for improved image-text alignment.
Second, we add an auxiliary token trained to predict the mini-mental state examination (MMSE) score, a widely used clinical measure of cognitive function that correlates with AD severity. This provides additional supervision for fine-tuning. 
Applying lightweight prompt tuning to both image and text modalities, our approach achieves state-of-the-art performance on ADNI with only 1,504 training MRIs, outperforming methods trained on 27,161 MRIs, and shows strong zero-shot generalization on OASIS-2 and AIBL. Code is available at \url{https://github.com/CFQ666312/DEFT-VLM-AD}.

\end{abstract}

\section{Introduction}
\label{sec:intro}

With the widely used vision-language models such as CLIP~\cite{radford2021learningCLIP} and BLIP~\cite{li2022blip}, AI systems have become increasingly capable of understanding both images and text by leveraging joint image-text representations. Following their success, medical vision language models~(Med-VLMs) have been proposed and achieved remarkable performance on classification~\cite{huang2021gloria}, segmentation~\cite{muller2022joint}, report generation~\cite{bruggeman2023collective} and many other tasks~\cite{shrestha2023medical,azad2023foundationalVLM}. 

Due to the significant gap between natural images and medical images, many Med-VLMs~\cite{wang2022medclip,eslami2023pubmedclip,zhang2023biomedclip,zhao2023clipbaseMVL} are pre-trained on large-scale datasets consisting of 2D medical images and text from open access papers~\cite{ROCO} or medical reports~\cite{MIMIC-CXR}. However, these models often require further adaptation for specific clinical applications. Currently, fine-tuning Med-VLMs for disease-specific tasks generally involves training both image and text encoders~\cite{ghosh2024mammo,felfeliyan2024application}, which requires substantial amounts of data and computing resources. Additionally, for conditions such as Alzheimer’s disease (AD), which rely on 3D MRI scans, existing models fail to fully exploit the spatial structure inherent in volumetric data. Therefore, developing a data-efficient fine-tuning approach based on pre-trained 3D Med-VLMs, such as those proposed in~\cite{bai2024m3d,shi2024med2e}, is crucial for accurate and clinically practical diagnosis of these diseases.


In this paper, we propose a data-efficient pipeline for fine-tuning 3D Med-VLMs to diagnose AD using 3D MRI data. 
A key challenge in AD diagnosis is the lack of detailed radiology reports. Given that the dataset contains metadata describing subject-specific information and MRI-derived attributes, we propose leveraging this information to generate synthetic reports, thereby providing richer textual supervision. 
Moreover, most existing models lack integration of clinical diagnostic knowledge, which is vital for informed decision-making in real-world settings. Since cognitive assessments, such as mini-mental state examination (MMSE) scores, play a crucial role in clinical diagnosis of AD, we introduce a learnable auxiliary token for predicting the MMSE score. This auxiliary prediction task further guides the tuning of image and text representations.
To address the domain gap arising from differences in imaging principles and characteristics between MRI and other radiological modalities, we adopt prompt tuning~\cite{jia2022vpt}, a lightweight and parameter-efficient fine-tuning approach that adapts both the image and text encoders. Additionally, a cross-attention mechanism is applied to enhance alignment between visual and textual modalities. 

We evaluate the proposed fine-tuning pipeline on three AD datasets, distinguishing AD, mild cognitive impairment (MCI), and normal controls (NC). Using only 1,504 MRIs for fine-tuning, our approach consistently improves classification performance across various pre-trained 3D Med-VLMs. On the ADNI dataset, our pipeline improves RadFM~\cite{wu2023towards} by 24.5\%, M3D by 23.9\% and Med3DVLM by 25.9\%. Notably, it achieves 78.8\% accuracy in this three-way classification task, outperforming MedBLIP~\cite{chen2024medblip} trained on 27,161 MRIs. Furthermore, our method outperforms other fine-tuning approaches under the same training conditions. Additionally, visualization results validate that textual input effectively guides the model to focus on clinically meaningful features, such as ventricular size. 

In short, our contributions are three-fold: 
\begin{enumerate}
    \item We build a data-efficient pipeline to fine-tune 3D Med-VLMs for 3D MRI, utilizing prompt learning for domain adaptation and cross-attention for modality alignment. 
    \item  We exploit metadata by generating synthetic reports, which provide stronger supervision to enhance the learning of visual features.
    \item  We add an auxiliary token to predict MMSE, incorporating clinical assessments into Med-VLMs.
\end{enumerate} 

\section{Related work}
\subsection{Medical vision language models}

Recent advances in VLMs have led to promising results in a variety of medical imaging tasks, including report generation, visual question answering (VQA), and disease classification. For example, MedCLIP~\cite{wang2022medclip} and BiomedCLIP~\cite{zhang2023biomedclip} typically leverage contrastive learning on paired image-text datasets to learn multimodal representations. While effective, most of these models are trained on large-scale 2D radiographs (e.g., chest X-rays or CT slices) and associated free-text reports, which inherently lack spatial context and anatomical continuity—both crucial for interpreting 3D medical scans like MRI and CT volumes.

Recently, several 3D vision-language models (VLMs) have been proposed for medical applications, utilizing different imaging modalities and backbone architectures. For instance, RadFM~\cite{wu2023towards} and M3D~\cite{bai2024m3d} adopt 3D  Vision Transformers~(ViTs) trained on 3D CT data, while Merlin~\cite{blankemeier2024merlin} employs a 3D ResNet as its image encoder. CT-CHAT~\cite{hamamci2024developing} leverages a specialized ViT variant, referred to as CT-ViT.  Furthermore, DCFormer~\cite{ates2025dcformer} proposes a dedicated encoder for 3D medical imaging, which enhances the extraction of spatial characteristics by factorizing standard 3D convolutions into three independent 1D convolutions applied separately along the depth, height, and width dimensions~\cite{xin2025med3dvlm}.

\subsection{Clinical knowledge integration in medical VLM}

While medical VLMs have achieved notable success in aligning visual information with free-text reports, such reports are not always available, especially in the case of certain datasets or disease types. Moreover, many existing models still lack integration of structured clinical diagnostic knowledge. To address theses limitations, prior studies have explored several alternatives. One line of work involves generating synthetic reports using large language models, such as GPT, or directly inserting human expert knowledge, as in~\cite{gao2024aligning}.
Other methods utilize structured clinical knowledge graphs derived from resources such as the Unified Medical Language System (UMLS)~\cite{wei2024integrating}. In addition, some approaches explicitly leverage prior knowledge by designing visual prompts, such as bounding boxes, circles, or masks highlighting important regions in medical images, to guide the learning process~\cite{denner2024visual}.

However, these methods face several limitations: synthetic reports may overlook patient-specific nuances; knowledge graphs are often difficult to integrate with visual data; and explicit visual prompts typically demand expensive manual annotation. Our approach addresses these challenges by converting metadata into corresponding reports and incorporating an auxiliary MMSE prediction task. This enables better alignment between 3D images and clinical knowledge, without relying heavily on free-text reports or manual prompts. 

\subsection{Efficient fine-tuning in low-data regimes}

The high cost of collecting and analyzing medical data presents a challenge for training VLMs. To address this, parameter-efficient fine-tuning~(PEFT) techniques have been proposed to adapt VLMs to specific tasks with minimal updates or even freeze the backbone model. Approaches such as prompt tuning~\cite{zhou2022learning, jia2022vpt}, adapter modules~\cite{gao2024clipadapter}, and low-rank adaptation (LoRA)~\cite{zhu2024melo} have demonstrated strong performance in natural language processing (NLP) and are now being actively investigated for medical imaging applications.

\section{Method}
\label{sec:Method}

\begin{figure*}[!t]
  \centering
\includegraphics[width=.7\textwidth,trim=0 280 0 60,clip]{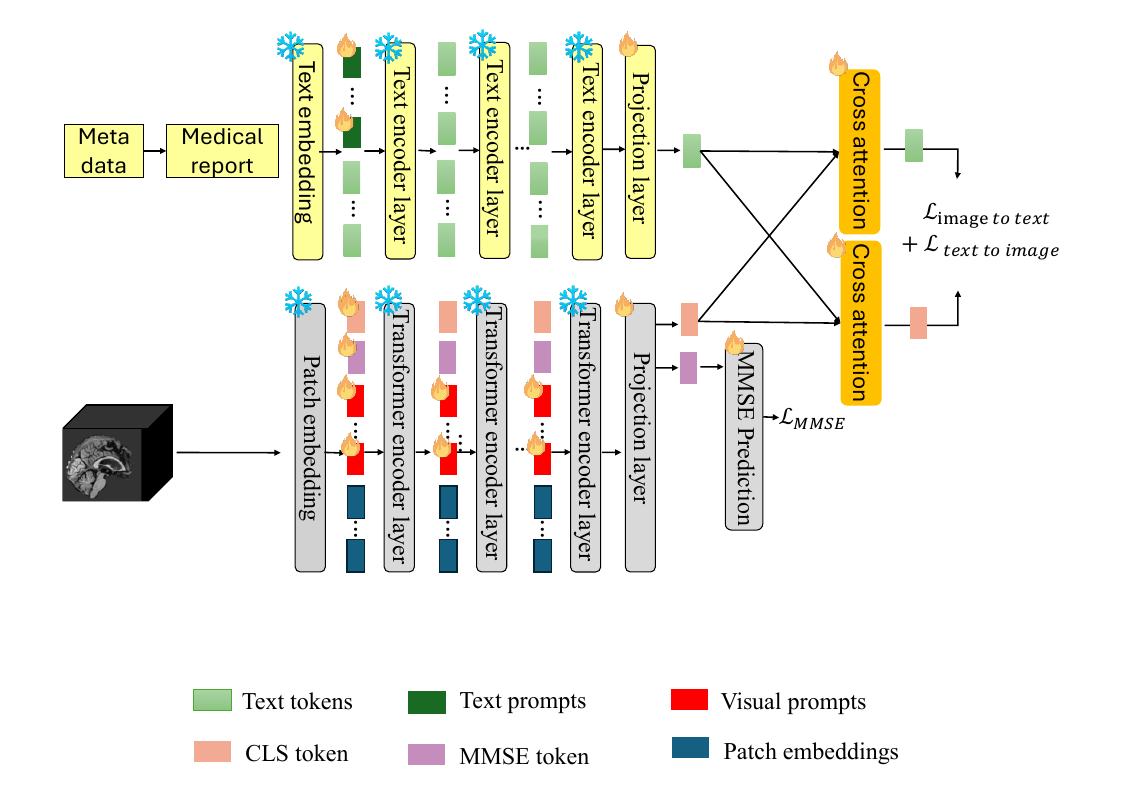}
\caption{Overview of the fine-tuning pipeline. The model is trained for image-text alignment and MMSE prediction. The upper part shows report processing with learnable prompts, while the lower part illustrates MRI processing with visual prompts and an auxiliary token. Cross-attention refines image-text feature alignment.} 
\label{framework}
\end{figure*}

Figure~\ref{framework} illustrates the proposed data-efficient fine-tuning pipeline. For the visual modality, we utilize the pre-trained 3D ViT as the vision encoder to process 3D medical images and incorporate visual prompt learning~\cite{jia2022vpt} for adapting from other modalities to MRI. An additional MMSE token is included alongside the standard patch embeddings and class token; it predicts the MMSE score, mimicking clinical diagnosis and guiding fine-tuning. For the text modality, we leverage the transformer-based text encoder~\cite{radford2021learningCLIP, bai2024m3d} to process synthetic reports generated from metadata and introduce learnable prompts to enhance its understanding of AD-related concepts. Prior to calculating image-text similarity, cross-attention is used to dynamically adjust the importance of attributes and improve alignment between the two modalities.

\subsection{Vision encoder}

Let $I \in \mathbb{R}^{D \times H \times W }$ denote the input 3D MRI, where $D,H,W$ denote the depth, height, and width, respectively. The input MRI is divided into $m$ 3D patches  $\{ I_n \in \mathbb{R}^{P_D * P_H * P_W} \mid n =1,2,\ldots,m\}$, where $P_D \times P_H \times P_W$ denotes the patch size. Each patch $I_n$ is encoded into a $d$-dimensional patch embedding $e_0^n$ by the patch embedding module:
\begin{equation}
e_0^n = \mathit{PatchEmbed}(I_n), \quad e_0^n \in \mathbb{R}^d, \quad n = 1, 2, \ldots, m,
\end{equation}
where the patch embedding module, $\mathit{PatchEmbed}$, consists of a 3D convolutional layer with kernel size and stride equal to the patch size, followed by a learnable positional embedding.
The set of patch embeddings is represented as $\mathbf{E}_0 = \{ e_{0}^n \in \mathbb{R}^{d} \mid n=1,2,\ldots,m\}$. These patch embeddings are concatenated with a learnable class token, $[\mathit{CLS}]$, and fed into the ViT. The ViT consists of $n_{\ell}$ layers, where the $\ell$-th transformer layer is denoted as $L_\ell$.

In addition to the class token, we introduce an auxiliary learnable token, $[\mathit{MMSE}]$, trained to predict MMSE score, where lower scores indicate more severe cognitive impairment. This auxiliary prediction provides additional supervision, encouraging the model to extract clinically relevant features from MRI scans.

As there is no publicly available 3D Med-VLM pre-trained on MRI data, we adopt a 3D ViT pre-trained on other imaging types, such as CT and X-ray, as the backbone vision encoder and apply deep visual prompt tuning (VPT)~\cite{jia2022vpt} to address the domain shift. VPT is a parameter-efficient fine-tuning technique which introduces a small number of learnable prompt tokens for model adaptation. Specifically, we incorporate learnable prompts $\mathbf{P}_{\ell}$ into each transformer layer ${\ell}$, with each prompt having a length of $l_p$:
\begin{equation}
\mathbf{P}_{\ell} = \{ P_{\ell}^k  \in \mathbb{R}^d \mid k=1,2,\ldots,l_p \}.
\end{equation}
These prompt tokens are concatenated with the standard ViT input sequence at each layer. The updated representation at layer ${\ell}$ is computed as:
\begin{equation}
[C_{\ell}, M_{\ell}, \__, \mathbf{E}_{\ell}] 
= L_{\ell}([C_{\ell-1}, M_{\ell-1}, \mathbf{P}_{\ell-1}, \mathbf{E}_{\ell-1}]), \\
\quad \ell = 1, 2, \ldots, n_{\ell},
\end{equation}
where $C_{\ell}$ and $M_{\ell}$ denote the embedding of the $[\mathit{CLS}]$ and $[\mathit{MMSE}]$ tokens respectively at layer $\ell$, $\mathbf{E}_{\ell}$ denotes the patch embeddings. After the final transformer layer, we extract the class and MMSE token embeddings $[C_{n_\ell}]$ and $[M_{n_\ell}]$. The class token embedding is then projected into a joint vision-language embedding space via a linear layer followed by   $l_2$ normalization, denotes as $\textit{f}_v$:
\begin{equation} 
\mathbf{v} = \textit{f}_v(C_{n_\ell}) .
\end{equation}
The resulting vector $\mathbf{v}$ will be used to compute image-text similarity.

For the final MMSE embedding $M_{n_\ell}$, we use a prediction head $\textit{f}_\mathit{mmse}$ composed of two linear layers with a non-linear activation between them to predict the MMSE score $\hat{y}_\mathit{mmse}$: 
\begin{equation}
\hat{y}_\mathit{mmse} = \textit{f}_\mathit{mmse} (M_{n_\ell}).
\end{equation}

\subsection{Text encoder}

A considerable amount of work focuses on radiology report generation, where models learn to produce clinical text from medical images or other large language models~\cite{pellegrini2023radialog, gu2025radalign}. 
These methods typically optimize language generation objectives to produce clinical reports. 
However, while they can indirectly improve image representations, they are not explicitly designed to extract discriminative features for diagnosis. 
In contrast, our approach aims to address the issue of missing MRI-related reports and to guide image representation learning for AD, without producing full reports in natural language. 
Therefore, we generate structured textual descriptions from MRI metadata. The report, denoted as~$T$, is created using attributes extracted from the metadata that reflect changes in brain structure during the progression of the disease. For example, the generated report might read: ``A photo of NC. The MRI scan reveals the following biomarkers: Hippocampal volume: 7323.00 $mm^3$, Ventricular size: 43767.00 $mm^3$, Whole brain volume: 968731.00 $mm^3$, Entorhinal cortex volume: 4056.00 $mm^3$, Fusiform gyrus volume: 18775.00 $mm^3$, Middle temporal gyrus volume: 17048.00 $mm^3$.'' 

During processing, $T$ is tokenized into a sequence of tokens, $\mathit{Tokens}$, using a subword tokenizer. Since the text encoder is pre-trained on natural language and may lack sufficient knowledge of AD, we incorporate the text prompt $\mathbf{R}$ of length $l_r$ at the input layer to enhance the text encoder $E_{text}$'s understanding of AD-related information:
\begin{equation}
\mathbf{R} = \{ R^q \mid q =1,2,3,\ldots,l_r \}.
\end{equation}
Therefore, the text features are obtained as:
\begin{equation}
\mathbf{w} = \textit{f}_t~(E_{text~}~([\mathbf{R}, \mathit{Tokens}])),
\end{equation}
where $\textit{f}_t$ is the linear projection layer followed by $l_2$ normalization.

\subsection{Attention-based aggregation}

To differentiate the significance of variables in the generated reports, we incorporate cross-attention layers following the projection layers for both image and text embeddings. This allows image features to integrate textual information while refining text features with visual information, dynamically adjusting the attributes importance for better alignment between MRI and reports. Following~\cite{vaswani2017attention}, we introduce a cross-attention layer to refine the features:
\begin{equation}
\hat{\mathbf{v}} = \text{softmax}\left( \frac{\mathbf{v} \cdot \mathbf{w}^\top}{\sqrt{d_k}} \right) \mathbf{w} + \mathbf{v},\ 
\hat{\mathbf{w}} = \text{softmax}\left( \frac{\mathbf{w} \cdot \mathbf{v}^\top}{\sqrt{d_k}} \right) \mathbf{v} + \mathbf{w},
\end{equation}
where $d_k$ is the dimensionality of the key and query vectors per attention head, used as the scaling factor~\cite{vaswani2017attention}.

\subsection{Loss functions}

For a batch of images with a batch size of $N$,  the similarity between image 
$i$ and text $j$ can be defined as:
\begin{equation}
s_{ij} = \frac{\hat{\mathbf{v}}_i \cdot \hat{\mathbf{w}}_j}{\|\hat{\mathbf{v}}_i\| \|\hat{\mathbf{w}}_j\|}.
\end{equation}
We expect image $i$ to have the highest similarity with its corresponding text, and likewise, text $j$ to have the highest similarity with its corresponding image, leading to the image-to-text (I2T) and text-to-image (T2I) contrastive losses: 
\begin{equation}
\small
\mathcal{L}_{\text{I2T}} = - \frac{1}{N} \sum_{i=1}^N \log \frac{\exp s_{ii}/\tau}{\sum_{j=1}^N \exp s_{ij}/\tau},\ \mathcal{L}_{\text{T2I}} = - \frac{1}{N} \sum_{j=1}^N \log \frac{\exp s_{jj} /\tau}{\sum_{i=1}^N \exp s_{ij}/ \tau},
\end{equation}
where $\tau$ is a learnable temperature parameter that controls the sharpness of the similarity distribution. 

For MMSE prediction, we calculate the mean squared error between the ground truth and the predicted value:
\begin{equation}
\mathcal{L}_\mathit{mmse} = \frac{1}{N} \sum_{i=1}^N \left( \hat{y}_\mathit{mmse}^{(i)} - y_\mathit{mmse}^{(i)} \right)^2.
\end{equation}

The final total loss is:
\begin{equation}
\mathcal{L} =  \mathcal{L}_{\text{I2T}} + \mathcal{L}_{\text{T2I}} + \lambda \mathcal{L}_\mathit{mmse} ,
\end{equation}
where $\lambda$ is a weighting coefficient that balances the relative importance of the contrastive loss and the MMSE prediction loss. 

During inference, the image is matched with the text prompt ``A photo of [class]'', and the class with the highest similarity score is assigned to the image.
\section{Experiment}

\subsection{Experiment setting}

\subsubsection{Datasets}

\begin{table*}
\centering
\small
\begin{tabular}{llcccc}
\toprule
Task & Dataset & \#Scans & \#Subjects & Age (min - max) & Gender (M/F) \\
\midrule
Fine-tuning   & ADNI  & 1,504  & 407 (107 NC, 134 MCI, 166 AD) & 54.4 - 90.9 & 214 / 193 \\
Classification   & ADNI & 268  & 79 (21 NC, 33 MCI, 25 AD) & 56.2 - 90.9 & 30 / 49 \\
Classification   & OASIS & 160  & 60 (30 NC, 30AD) & 60.0 - 93.0 & 23 / 37 \\
\bottomrule
\end{tabular}
\caption{Demographic information of the datasets used for fine-tuning and classification tasks.}
\label{tab:dataset}
\end{table*}

In our fine-tuning process, we use the public
Alzheimer’s Disease Neuroimaging Initiative (ADNI) dataset~\cite{ADNI}. A set of 1,504 3D T1-weighted MRI scans from 407 subjects is used to fine-tune Med-VLMs. These subjects include 107 normal controls (NC), 134 individuals with mild cognitive impairment (MCI) and 166 diagnosed with AD. Further demographic information is summarized in Table~\ref{tab:dataset}. 
Each MRI scan undergoes the following preprocessing steps: denoising, bias field correction, skull stripping, affine registration to a standard template, rescaling to a 128$\times$128$\times$128 volume, and normalization to the [0, 1] range~\cite{ouyang2021self}. 
As for MMSE scores in this dataset, the values range from 2 to 30. 

During evaluation, we assess all methods on three AD datasets. First, we evaluate on the ADNI dataset using 268 MRI scans from 79 subjects who are not included in fine-tuning, comprising 21 NC, 33 MCI, and 25 AD. Second, to assess the model's zero-shot generalizability on an unseen dataset, we evaluate on a balanced subset of the OASIS-2 dataset~\cite{marcusoasis}, which contains 160 MRIs from 60 subjects (30 demented, 30 non-demented), and on a subset of the AIBL dataset~\cite{fowler2021fifteen}, which contains 150 MRIs (50 NC, 50 MCI, and 50 dementia).

\subsubsection{Implementation details}

In our method, the text encoder is fixed and consists of a 12-layer transformer with 768 hidden dimensions, initialized from the pretrained model in~\cite{bai2024m3d}, while for the vision encoder, we considered three different 3D ViTs pre-trained on various tasks and datasets, namely M3D~\cite{bai2024m3d}, RadFM~\cite{wu2023towards}, and Med3DVLM~\cite{xin2025med3dvlm}. The patch size is 4 $\times$ 16 $\times$ 16. The dimension of image and text embeddings is 768. The visual prompt length is 20, and the text prompt length is 30. $\lambda$ is set to 0.5, and the learnable temperature parameter $\tau$ is initialized to 10.
The fine-tuning process uses the AdamW optimizer, with a learning rate scheduler of ReduceLROnPlateau, wherein the learning rate is reduced by a factor of 10 when the validation performance stagnates for 5 consecutive epochs. The fine-tuning is conducted using 2 NVIDIA Quadro RTX 8000 GPUs with 48 GB of memory each.

\subsubsection{Baseline methods}

To comprehensively evaluate the effectiveness of our pipeline, we conduct experiments in three progressive stages: (1) zero-shot evaluation, (2) performance improvement via fine-tuning, and (3) comparison with alternative fine-tuning methods. 

We categorize baseline MedVLMs into two groups: 2D-based and 3D-based models. In our experiments, we utilize their pre-trained encoders as backbones for evaluation.

For 2D-based MedVLMs, MedCLIP~\cite{wang2022medclip} is one of the representative models, trained primarily on chest X-ray datasets using contrastive learning. BioMedCLIP~\cite{zhang2023biomedclip} extends this approach by incorporating more diverse datasets, including multiview chest X-rays and other imaging types.

Regarding 3D MedVLMs, M3D~\cite{bai2024m3d} is a 3D MedVLM trained on CT data, leveraging a vision encoder pre-trained using contrastive image-text retrieval within a CLIP-based framework. RadFM~\cite{wu2023towards} is designed specifically for multimodal radiologic tasks and adopts a two-stage training procedure: large-scale pre-training on the Medical Multimodal Dataset (MedMD), followed by fine-tuning on the Radiology Multimodal Dataset (RadMD), both curated by the authors to facilitate robust cross-modal alignment and domain adaptation. The pretrained encoders from M3D and RadFM are directly evaluated on our dataset. Med3DVLM~\cite{xin2025med3dvlm} is pretrained on CT scans; it has been evaluated both via full-parameter training and fine-tuning on our AD dataset. All of these baselines utilize the ViT as the vision encoder. 

We also include MedBLIP~\cite{chen2024medblip}, a model originally developed as a computer-aided diagnosis (CAD) system for Alzheimer's disease, which can be responsible for the classification task. Since MedBLIP performs suboptimally when directly trained on our dataset, we report its results from original paper as a dedicated baseline for comparison.

To further assess the effectiveness of pipeline, we compare with other fine tuning approaches. The method proposed by Bai et al.~\cite{bai2024m3d} introduces a 3D spatial pooling module to reduce the dimensionality of 3D image features, reducing computational costs while preserving spatial information. This module is combined with LoRA-based fine-tuning. Furthermore, LGA~\cite{hu2024lga} addresses domain adaptation challenges by integrating a cross-modal adapter module.

\subsection{Quantitative result of classification task}

\begin{table}[t]
\caption{Classification accuracy of different backbones with and without fine-tuning on the ADNI, OASIS and AIBL datasets, as well as computational cost. 
Our results are averaged over five random runs and reported as mean $\pm$ standard deviation. 
$\dagger$ indicates results taken from the original paper.}
\label{ViT}
\centering
\setlength{\tabcolsep}{2pt}
\begin{tabular}{lcccccc}
\toprule
\textbf{Methods} & \multicolumn{3}{c}{\textbf{ACC (\%)}} & \# Parameters &  Time & Mem.\\
 & ADNI & OASIS & AIBL & (Learnable) & /Epoch & /Card\\
 \midrule
\textbf{MedCLIP~\cite{wang2022medclip}} & 39.6 & 40.0 & 34.7 &200M~(0) & - & -\\
\midrule
\textbf{BioMedCLIP~\cite{zhang2023biomedclip}} & 40.7 & 41.3 & 25.3 & 210M~(0) & - & -\\
\midrule
\textbf{RadFM~\cite{wu2023towards}} & 52.9 & 60.4 & 35.3 & 120M~(0) &- &-\\
+ 3D SP Perceiver~\cite{bai2024m3d} & 69.8 & 68.1 & 51.3 & + 35.7M & 508.6s & 43.4G\\ 
+ LGA~\cite{hu2024lga} & 74.6 & 76.3 & 57.3 & + 85.2M & 932.4s & 46.1G\\
+ Ours & \textbf{77.4$\pm$2.0} & \textbf{78.1$\pm$1.8} &  \textbf{62.8$\pm$5.5} & + 12.2M & 248.2s & 30.9G \\
\midrule
\textbf{M3D~\cite{bai2024m3d}} & 54.9 & 52.6 & 34.0 & 198M~(0) & -&-\\
+ 3D SP Perceiver~\cite{bai2024m3d} & 74.3 & 78.8 & 54.7 & + 35.7M & 508.6s & 43.4G\\ 
+ LGA~\cite{hu2024lga} & 76.7 & 80.0 & 52.0 & + 85.2M &  932.4s & 46.1G\\
+ Ours & \textbf{78.8$\pm$3.2} & \textbf{81.6$\pm$3.7} & \textbf{71.3$\pm$4.2} & + 12.2M & 248.2s & 30.9G\\
\midrule
\textbf{Med3DVLM~\cite{xin2025med3dvlm}} & 50.4 & 48.3 & 44.7 & 199M~(0) &- &-\\
+ 3D SP Perceive~\cite{bai2024m3d} &74.6& 79.4 & 68.0 & + 35.7M & 508.6s & 43.4G \\ 
+ LGA~\cite{hu2024lga} & 75.0 & 78.1 & 67.3 & + 85.2M & 932.4s & 46.1G\\
+ Ours & \textbf{76.3$\pm$1.9} & \textbf{79.7$\pm$3.5} & \textbf{69.1$\pm$3.3} & + 12.2M & 248.2s & 30.9G\\
\midrule
MedBLIP~\cite{chen2024medblip}${}^\dagger$ &  78.7 & 85.3 & 80.8 & 1.7B~(154M) &- &-\\
\bottomrule
\end{tabular}
\end{table}

We first evaluate the zero-shot performance of existing MedVLMs that were not pre-trained on AD-related data. As shown in Table~\ref{ViT}, 2D-based MedVLMs such as MedCLIP~\cite{wang2022medclip} and BioMedCLIP~\cite{zhang2023biomedclip}, which are mainly trained on chest X-rays and other 2D medical images, perform poorly when directly applied to our 3D brain MRI-based classification task. Their limited accuracy highlights the difficulty of cross-modal transfer between 2D and 3D data, especially for structure-specific diseases like AD. 

3D-based models such as M3D~\cite{bai2024m3d} and RadFM~\cite{wu2023towards}, which are pre-trained on CT or multimodal radiologic datasets, perform better than 2D models but still fall short in our AD task. This performance gap emphasizes the importance of both modality alignment and task specificity. Among these, Med3DVLM~\cite{xin2025med3dvlm}, which is trained on our dataset, outperforms others by at least 10\%. This suggests that domain adaption during pre-training is crucial for downstream classification accuracy.

We next evaluate how much we can improve the poor zero-shot performance of backbones not trained on the corresponding image modality, while demonstrating that fine-tuning still outperforms full training the encoder on our dataset. Our results show that fine-tuning M3D and RadFM with our pipeline improves their accuracy by at least 23\%, indicating that our approach can significantly alleviate the limitations imposed by domain mismatch and lack of task-specific information. The comparison between full training Med3DVLM and fine-tuning demonstrates that our approach is both efficient and effective in adapting pre-trained models to disease-specific tasks.   

To further contextualize our performance, our method achieves 78.8\% accuracy, surpassing MedBLIP~\cite{chen2024medblip}, which was specifically designed for AD diagnosis and trained on a large-scale MRI dataset. Notably, our approach reaches this performance with significantly fewer trainable parameters, demonstrating both high efficiency and strong task adaptability.

Beyond improving zero-shot and outperforming full training baselines, our method is further compared with other fine-tuning approaches to assess its relative effectiveness. As shown in Table~\ref{ViT}, compared to other fine-tuning methods, our pipeline further highlights the superiority of leveraging pre-trained knowledge for accurate AD classification.

Finally, we evaluate zero-shot performance on the OASIS-2 and AIBL datasets. Our pipeline outperforms other fine-tuning strategies and achieves strong zero-shot results without additional fine-tuning. MedBLIP reports high performance on both datasets. However, it should be noted that this method was pre-trained on 27,161 MRIs from ADNI, NACC, and OASIS, making the OASIS evaluation not truly zero-shot.

\subsection{Ablation studies}

\subsubsection{Impact of individual component}

\begin{table*}
\centering
\setlength\tabcolsep{2pt}
\begin{tabular}{cccccccc}
\toprule
& \multicolumn{2}{c}{Image Encoder}  & \multicolumn{2}{c}{Text Encoder} & Cross-Attention &{ACC} \\ 
& Visual prompts  & MMSE token & Textual prompts & Report & & \\
\midrule
(a) & &   &  & & &54.9\%\\
\midrule
\multirow{3}{*}{(b)} & \checkmark &  & & & &65.3\%\\
 &  & \checkmark & & & &67.1\%\\
& \checkmark & \checkmark &  & & & 68.9\% \\
\midrule
\multirow{3}{*}{(c)} &  &  & \checkmark & & &63.6\% \\
 &&  &  & \checkmark & &  74.4\% \\
 &&  & \checkmark & \checkmark & &75.6\% \\
\midrule
(d) & \checkmark & \checkmark & \checkmark & \checkmark & & 77.8\% \\
\midrule
(e) & \checkmark & \checkmark & \checkmark & \checkmark &\checkmark & 78.8\% \\
\bottomrule
\end{tabular}
\caption{Evaluation of the different components of the proposed method.}\label{ablation}
\end{table*}

We assess the effectiveness of each component in our fine-tuning pipeline on ADNI using M3D as the backbone (Table~\ref{ablation}). Compared to the baseline (a), fine-tuning the image encoder (b) or text encoder (c) improves accuracy, with the highest achieved when both are fine-tuned (d). Among components, generated reports provide the largest gain, highlighting the benefit of structured metadata, while the MMSE prediction token also contributes to capturing disease-specific patterns. These results demonstrate the complementary contributions of different components.

Given the impact of generated reports in (c), we further examine metadata. Using the fine-tuned model (d), we extract image features from the vision encoder and numeric biomarker features from the text encoder, and train a simple linear classifier on each type and their concatenation to evaluate predictive value independently of the full pipeline. All classifiers are trained and tested on ADNI. As shown in Table~\ref{tab:classifier_comparison}, biomarkers alone effectively support classification, while combining image and biomarker features further improves performance.

\begin{table*}[t]
\centering
\begin{minipage}{0.48\textwidth}
\centering
\caption{Classification accuracy using different input features.}
\label{tab:classifier_comparison}
\begin{tabular}{lc}
\toprule
Input features & ACC (\%) \\
\midrule
image features & 63.9$\pm$3.9 \\
biomarker vectors & 65.9$\pm$3.8 \\
concatenated features & 72.0$\pm$4.1 \\
\bottomrule
\end{tabular}
\end{minipage}
\hspace{0.02\textwidth} 
\begin{minipage}{0.48\textwidth}
\centering
\caption{Classification accuracy under varying fine-tuning sample size on ADNI.}
\label{data_scale_results}
\begin{tabular}{lcccc}
\toprule
Sample size & 500 & 1000 & 1500 & 2000 \\
\midrule
ACC & 63.1\% & 66.8\% & 78.8\% & 81.3\% \\
\bottomrule
\end{tabular}
\end{minipage}
\end{table*}

\subsubsection{Impact of fine-tuning sample size}

To evaluate the data efficiency of our fine-tuning pipeline, we conduct experiments on subsets of the ADNI dataset with 500, 1000, 1500, and 2000 MRIs. As shown in Table~\ref{data_scale_results}, classification accuracy improves steadily with increasing data size. Notably, our method achieves strong performance with just 1500 samples, surpassing results obtained with fewer data. This indicates that our approach is capable of effective fine-tuning with limited annotated data, which is crucial for practical clinical scenarios where data collection can be challenging.

\subsubsection{Impact of fine-tuning strategies}

\begin{table}
\centering
\begin{tabular}{lccc}
\toprule
 &  prompt learning & LoRA & adapter\\
 \midrule
 ACC &78.8\% & 78.5\% & 75.0\% \\
\bottomrule
\end{tabular}
\caption{Comparison of different PEFT methods on ADNI dataset}\label{domain adaption}
\end{table}

To address the domain gap issue in medical imaging tasks under limited supervision, we adapt prompt tuning in our vision encoder, which is demonstrated effective in Table~\ref{ablation} $(b)$. In this section, we compare it with other popular parameter-efficient fine-tuning (PEFT) methods, such as Adapter~\cite{gao2024clipadapter} and LoRA~\cite{zhu2024melo}, to validate the effectiveness of our framework and demonstrate the flexibility of our pipeline. 

As shown in Table~\ref{domain adaption}, prompt learning selected by our pipeline yields strong results, particularly due to its ability to dynamically adjust task-specific prompts at each layer without modifying the backbone. Importantly, our pipeline is flexible and modular, allowing easy integration of various PEFT strategies such as prompt tuning, LoRA, or Adapters. This flexibility enables practitioners to select the most suitable fine-tuning method depending on deployment constraints, model accessibility, or domain-specific requirements. In our experiments, prompt learning showed the best overall performance, but our framework supports switching to other methods when appropriate. 

\subsubsection{Interpreting biomarker importance}

\begin{figure}[t]
    \centering
    \includegraphics[width=\textwidth]{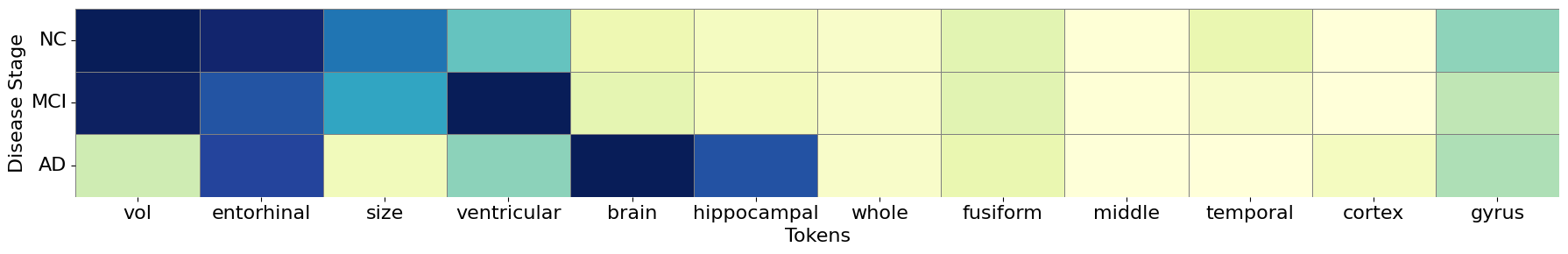}
    \caption{Heatmap of integrated gradients showing the importance of each biomarker for different disease progression stages~(darker color indicates higher importance).}
    \label{fig:biomarker_heatmap}
\end{figure}

To understand how our model leverages textual features for image-text alignment, we use Integrated Gradients~\cite{sundararajan2017axiomatic} to attribute contributions of individual text tokens to matching scores. As shown in Figure~\ref{fig:biomarker_heatmap}, for a single subject across three disease stages, the top contributing words shift: early-stage tokens like ``vol'', ``entorhinal'', and ``size'' highlight volumetric measures; intermediate-stage tokens such as ``entorhinal'' and ``ventricular'' reflect advancing structural changes; and late-stage tokens like ``brain'' and ``hippocampal'' indicate focus on atrophy or progression. These results show that the model’s decisions align with key clinical biomarkers, supporting its relevance in disease assessment.

\subsubsection{Visualization analysis}

\begin{figure}
\centering
\includegraphics[width=\linewidth]{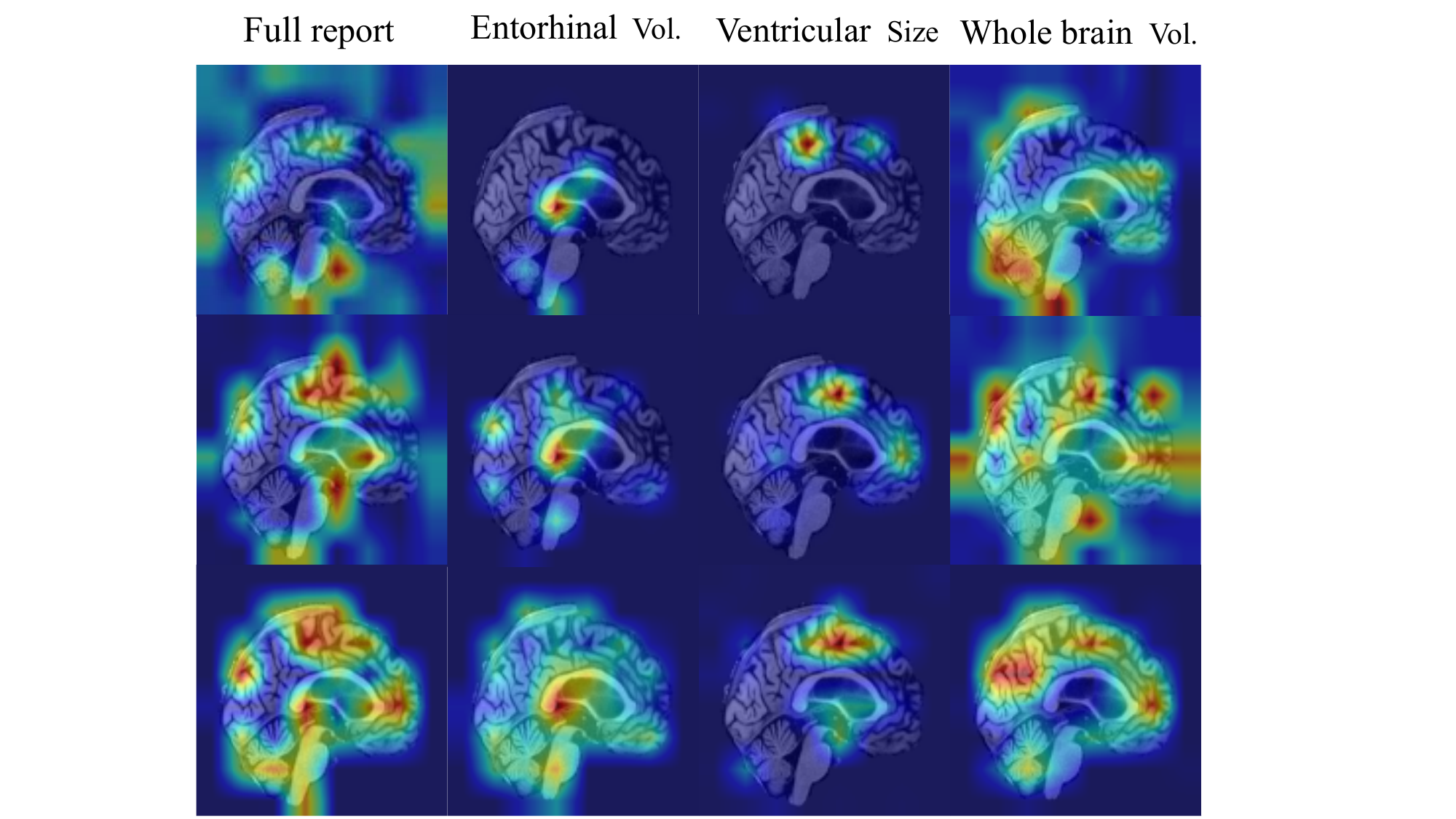}
\caption{Visualization of Grad-ECLIP attention heatmaps generated using different textual inputs, illustrating the individual contributions of Entorhinal Volume, Ventricular Size, and Whole Brain Volume to image-text alignment. The warmer parts of the heatmaps indicate regions where the model places greater importance when matching the image with the given biomarker text.Each row corresponds to MRI scans of the same subject at different clinical stages: NC,MCI,AD}
\label{gradcam}
\end{figure}

In addition to analyzing the contribution of individual text tokens, we further investigate the spatial correspondence between specific biomarkers and image regions. To achieve this, we utilize Grad-ECLIP~\cite{zhao2025grad}, a gradient-based interpretability method that generates heatmaps showing how much each image patch contributes to the matching score~\cite{zhao2025grad} conditioned on a selected biomarker token. For each experiment, we retain only one biomarker in the textual input while masking the others. We then compute the gradient of the similarity score with respect to the image patch embeddings to generate a heatmap.

As shown in Figure~\ref{gradcam}, the heatmaps generated under different textual inputs reveal distinct activation patterns. When using entorhinal volume or ventricular size alone, the attention maps exhibit more localized and concentrated responses, primarily focusing on brain regions known to be affected in AD. These patterns are not only narrower in scope compared to the full report, but also show better alignment with known AD pathology, suggesting that the model has learned clinically meaningful associations.

When using whole brain volume as the only input, the attention tends to spread across broader brain areas. This suggests that whole brain volume may introduce less specific guidance for localization, and the model's attention becomes more diffuse. Overall, these results highlight that certain biomarkers -- particularly entorhinal volume and ventricular size -- are more informative and effective in guiding the model toward disease-relevant visual features.

\section{Conclusion}
In conclusion, this work presents a pipeline for adapting pre-trained VLMs to disease-specific tasks in data-constrained medical imaging applications. It achieves effective domain adaptation, shows the potential value of medical metadata, and supports auxiliary prediction. The fine-tuning strategies employed in our pipeline are orthogonal to the specific design for AD, indicating that more advanced techniques could replace prompt tuning to further improve performance. Future work could further enhance task generalizability by integrating richer medical knowledge, such as knowledge graphs, and improve interpretability by leveraging retrieval-augmented generation.

\section*{Acknowledgment}
Data for this project was funded by the Alzheimer's Disease Neuroimaging Initiative (ADNI) (National Institutes of Health Grant U01 AG024904). OASIS-2 data was obtained from the Open Access Series of Imaging Studies (OASIS) project, and AIBL data was obtained from the Australian Imaging, Biomarkers and Lifestyle (AIBL) study.

\bibliographystyle{ieeetr}
\bibliography{main}

\end{document}